\documentclass[conference]{IEEEtran}
\IEEEoverridecommandlockouts
\usepackage{cite}
\usepackage{amsmath,amssymb,amsfonts}
\usepackage{algorithmic}
\usepackage{graphicx}
\usepackage{textcomp}
\usepackage{xcolor}

\usepackage{epsfig}
\usepackage{balance}

\usepackage{url}
\usepackage{caption}
\usepackage{lipsum}
\usepackage{subfigure}
\usepackage{subfig}
\usepackage{color}
\usepackage{balance}
\usepackage{float}
\usepackage{multirow}
\usepackage{float}
\usepackage{breqn}
\usepackage{tabularx}

\usepackage{amsmath,url}

\usepackage{amsmath,amssymb,amsfonts}
\usepackage{algorithmic}
\usepackage[]{graphicx}
\graphicspath{{Figures/}}
\usepackage{subfig}
\usepackage{lipsum}
\usepackage{balance}
\usepackage{textcomp}
\usepackage{xcolor}
\usepackage{multirow}
\usepackage{color}
\usepackage{xcolor}
\usepackage{algorithm} 
\usepackage{algorithmic}
\usepackage{listings}
\usepackage{courier}
\usepackage{float}

\def\BibTeX{{\rm B\kern-.05em{\sc i\kern-.025em b}\kern-.08em
    T\kern-.1667em\lower.7ex\hbox{E}\kern-.125emX}}

\begin{document}

\title{Identifying Candidate Spaces \\for Advert Implantation}

\author{
Soumyabrata~Dev$^{1,2}$,
Hossein~Javidnia$^{1}$,
Murhaf~Hossari$^{1}$,
Matthew~Nicholson$^{1}$,
Killian~McCabe$^{1}$,\\
Atul~Nautiyal$^{1}$,
Clare~Conran$^{1}$,
Jian~Tang$^{3}$,
Wei~Xu$^{3}$,
and~Fran\c{c}ois Piti\'e$^{1,4}$
\\
$^{1}$ADAPT SFI Research Centre, Dublin, Ireland\\
$^{2}$School of Computer Science, University College Dublin, Dublin, Ireland\\
$^{3}$Huawei Ireland Research Center, Dublin, Ireland\\
$^{4}$Department of Electronic \& Electrical Engineering, Trinity College Dublin

\thanks{The  ADAPT  Centre  for  Digital  Content  Technology  is  funded  under  the  SFI Research Centres Programme (Grant 13/RC/2106) and is co-funded under the European Regional Development Fund.}
\thanks{
Send correspondence to F.\ Piti\'e, E-mail: PITIEF@tcd.ie.
}
}

\maketitle

\begin{abstract}
Virtual advertising is an important and promising feature in the area of online advertising. It involves integrating adverts onto live or recorded videos for product placements and targeted advertisements. Such integration of adverts is primarily done by video editors in the post-production stage, which is cumbersome and time-consuming. Therefore, it is important to automatically identify candidate spaces in a video frame, wherein new adverts can be implanted. The candidate space should match the scene perspective, and also have a high quality of experience according to human subjective judgment. In this paper, we propose the use of a bespoke neural net that can assist the video editors in identifying candidate spaces. We benchmark our approach against several deep-learning architectures on a large-scale image dataset of candidate spaces of outdoor scenes. Our work is the first of its kind in this area of multimedia and augmented reality applications, and achieves the best results. 
\end{abstract}


\section{Introduction}
\label{sec:intro}

Digital technology is used to insert a virtual image of a specific advert into a pre-recorded television show or a live sporting event. This technique enables advertisers to target specific audience and generate personalised recommendation to the consumers. The advertisement should be integrated seamlessly into the video frame without hampering the quality of the viewing experience of the viewers. Additionally, 
such technologies can assist the advertisers in generating personalized video content, that fits a targeted language, region, or age group. 

\begin{figure}[htb]
  \begin{center}
    \includegraphics[width=0.2\textwidth]{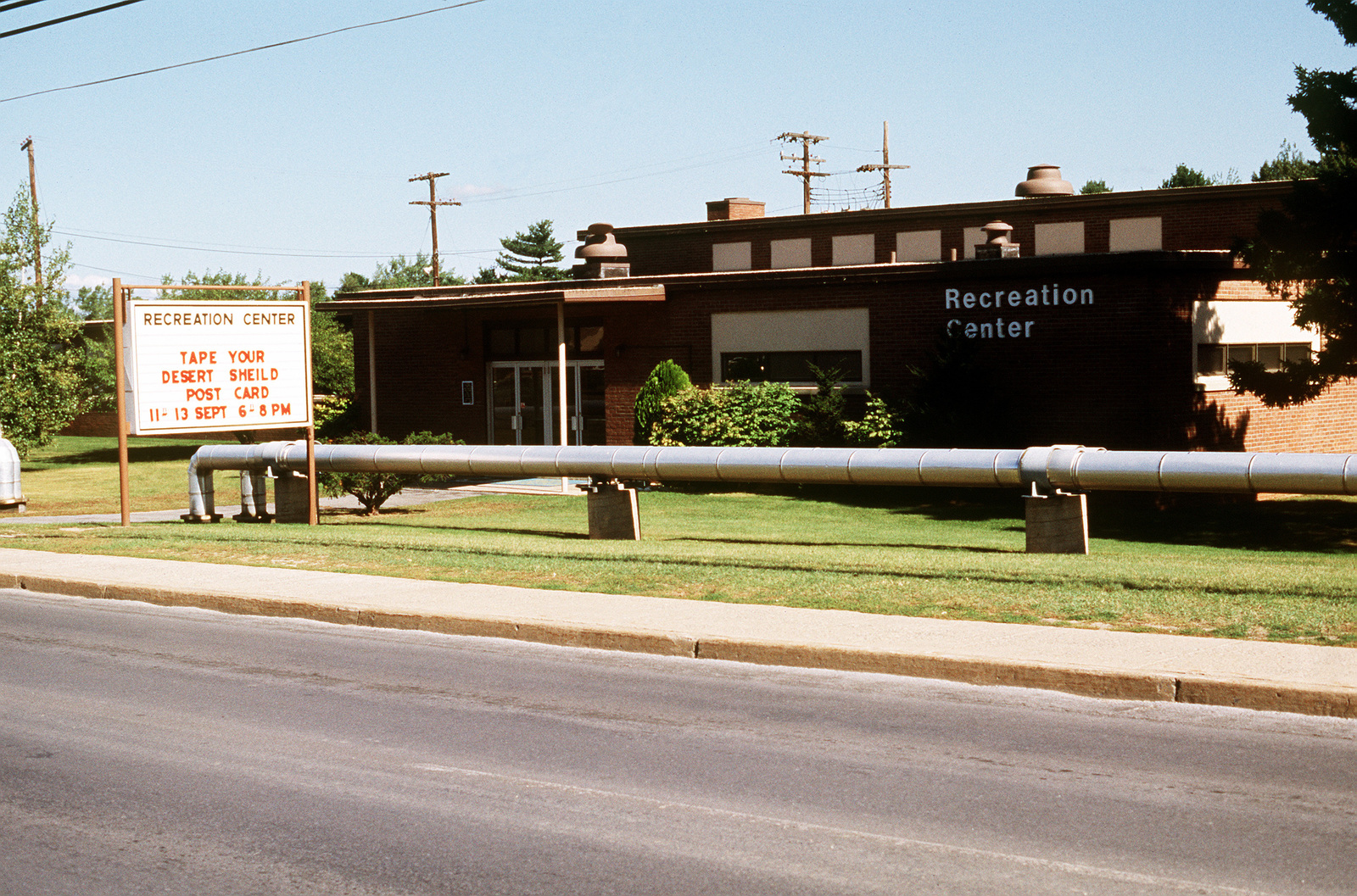}
    \includegraphics[width=0.2\textwidth]{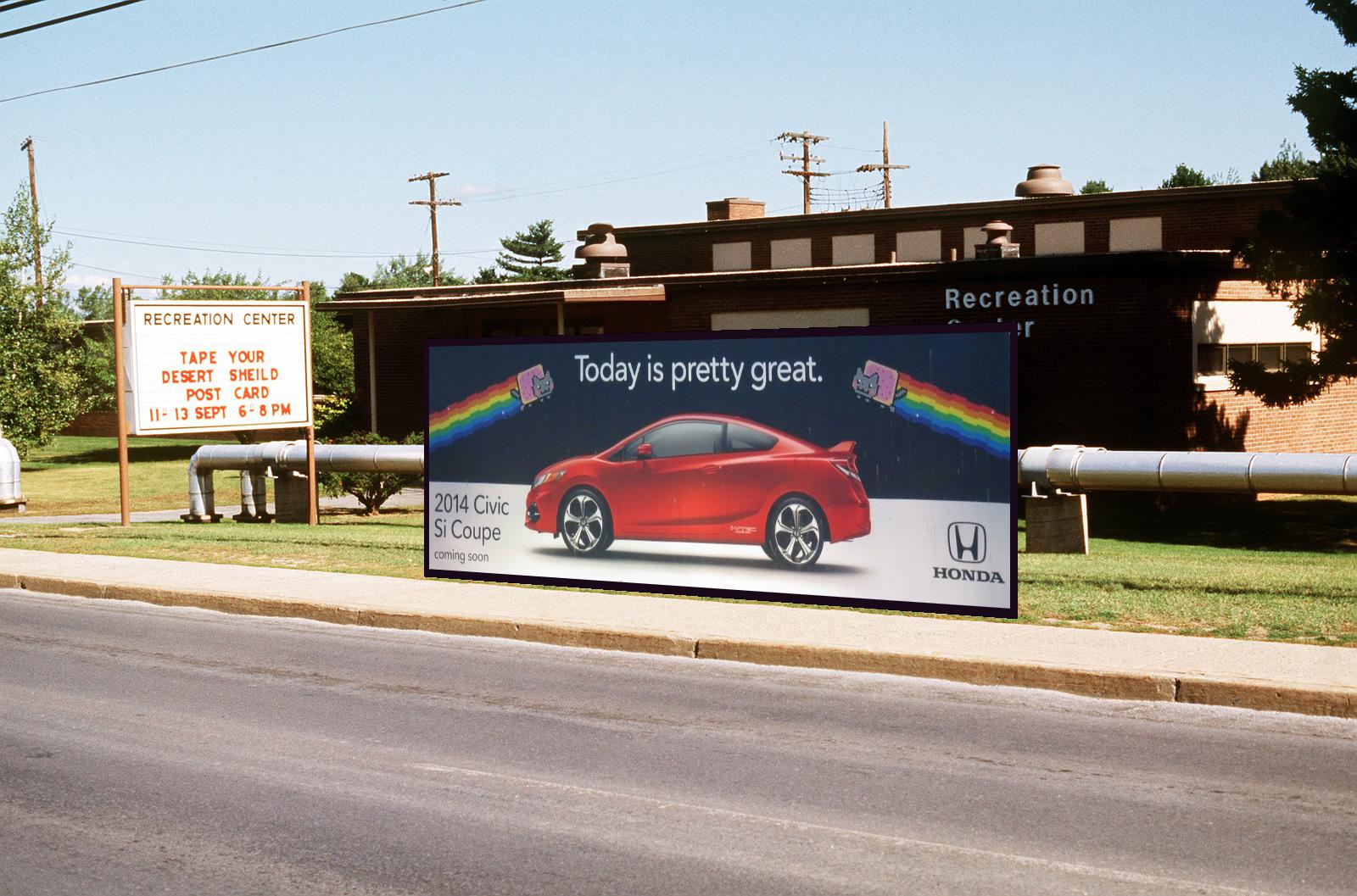}\\
    \makebox[0.45\textwidth][c]{(a) Advert implantation on a floating candidate space}\\
    \includegraphics[width=0.2\textwidth]{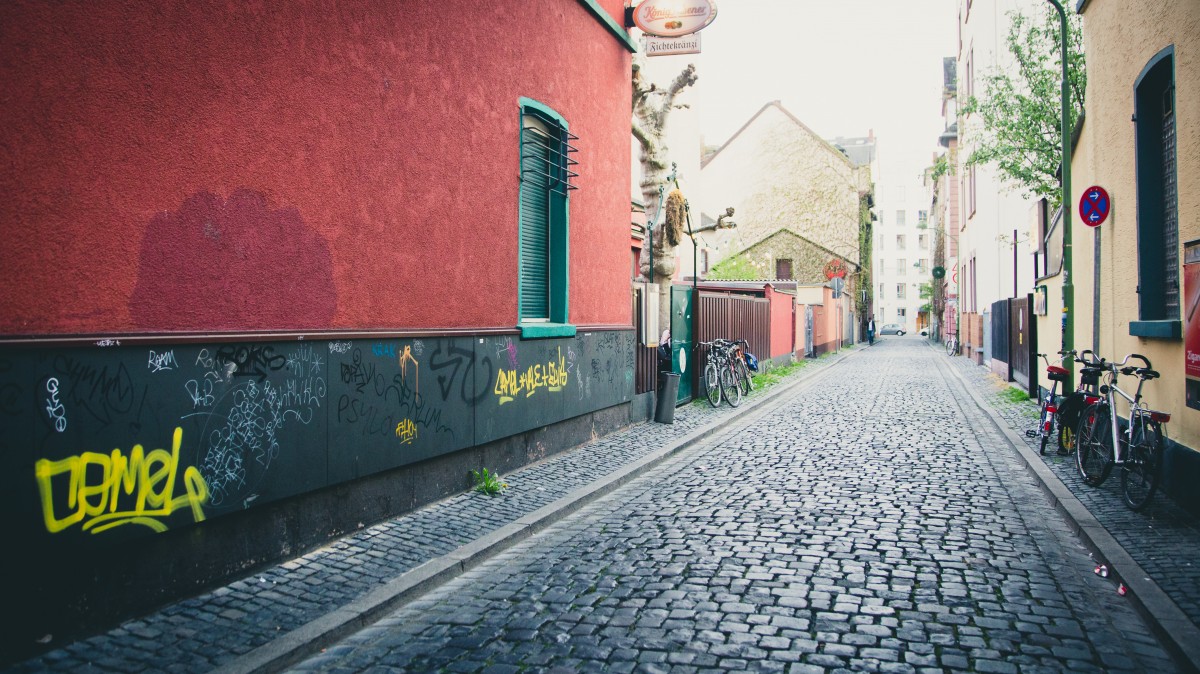}
    \includegraphics[width=0.2\textwidth]{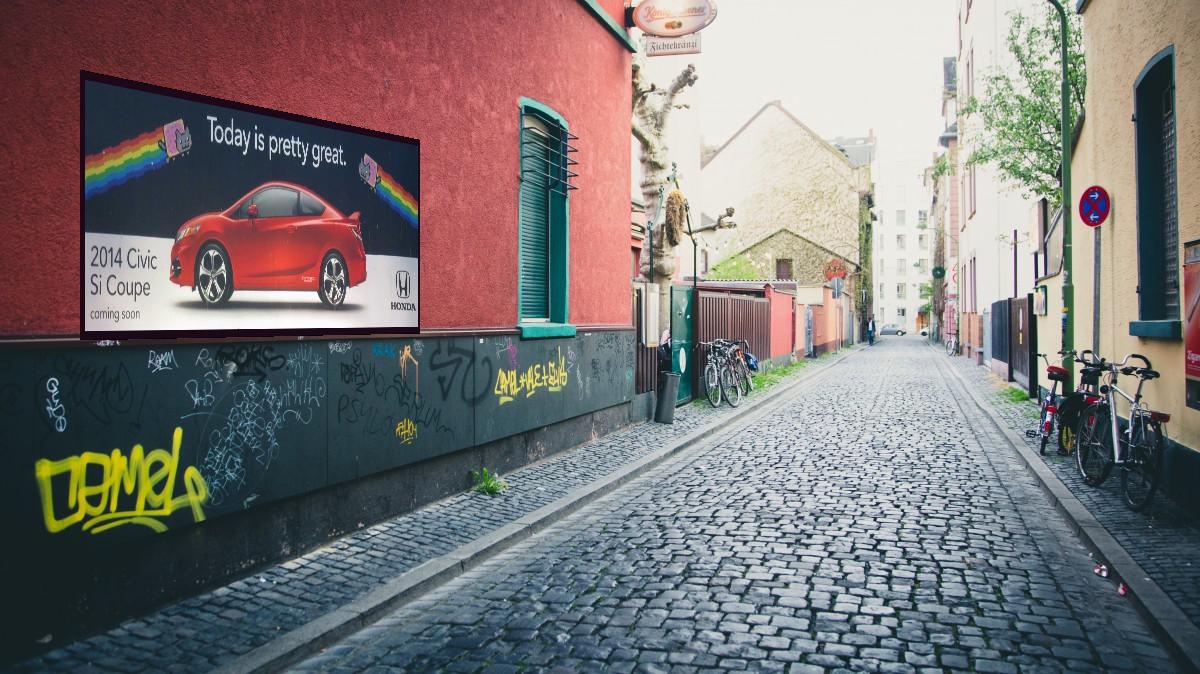}\\
    \makebox[0.45\textwidth][c]{(b) Advert implantation against an anchored candidate space}\\
  \end{center}
  \caption{Illustration depicting an advertisement implantation on (a) floating candidate space, and (b) anchored candidate space. In this paper, we attempt to identify such candidate spaces in outdoor scenes.}
  \label{fig:story}
\end{figure}

The ability to automatically detect potential spaces where advertisements can be placed, without disturbing the viewer or the flow of the scene is quite vital. Such capabilities of inserting advertisements~\cite{hossari2018adnet, dev2019localizing} in both recorded and live videos create great opportunities for both broadcasters and advertisers. Imagine you are watching your favorite television show, and during a specific scene of the show, you see a billboard advertisement of your favorite car or smartphone, by the side of the road. On a similar note, your friend residing half way across the globe watching the same television show visualises a different advertisement of his/her favorite music band that is scheduled to take place in the town of his/her residence. Such targeted advertisement avenues open up the potential of generating personalised video content to different demographics. Figure~\ref{fig:story} describes such a scenario wherein new advert is integrated into the original scene. The new advertisement can be integrated on a floating candidate space as described in Fig.~\ref{fig:story}(a). The advert can also be anchored against a fixed surface eg.\ wall, as shown in Fig.~\ref{fig:story}(b).

The main contributions of this paper are as follows: (a) we propose a methodology in identifying candidate spaces in recorded videos, where advertising images can be implanted without disturbing the viewing experience of the viewers, (b) we also propose a bespoke neural net that can automatically parse the video frames, and identify the \emph{appropriate} candidate spaces for advert integration. 
The candidate space should be appropriate in a manner that the advert is integrated in a realistic location of the scene, and the augmented video looks natural to the viewers. Our method uses deep-learning based convolutional neural network in order to process images or video frames, and proposes a probabilistic map of potential locations, which are suitable to insert an advertisement in. Our network is a light-weight network which is trained on a large corpus of image dataset with manually annotated candidate spaces, ensuring the best subjective judgments of the annotators. Using an encoder-decoder structure, our network is able to accurately identify possible locations on video streams or images where a localised virtual advertisement can be placed (candidate space). 
We have also compared our approach with different state-of-the-art architectures that can be used for similar tasks. Our technique outperformed the other architectures for the task of identifying the advertising candidate spaces, and being a light-weight architecture has the potential for faster training and quicker prediction. Our work is the first in this area and we show that our bespoke solution achieves competitive results in this specific domain.

\section{Advertisement Integration}

\subsection{Related work}
With the advent of high computing power and the existence of plethora of user-generated online videos, there are exciting developments in the realm of multimedia and advertising. In 2015, Stephan filed a US patent~\cite{stephan2015virtual} on virtual advertising platform that uses a mapping algorithm to insert an advertisement in video streams. MirriAd Ltd.\ has also filed several patents~\cite{popkiewicz2008process,popkiewicz2015apparatus} in this area of product placement and targeted advertisements. In our previous work~\cite{nautiyal2018advert}, we proposed an end-to-end implementation of a deep learning based system that can generate augmented video streams containing personalised advertisements. 
In the literature, Covell et al.\ uses a combination of audio and visual features to detect advertisement frames from the sequence of video frames~\cite{covell2006advertisement}. Similar work was reported in \cite{feng2013real}, where Feng and Neumann exploited audio-visual features with logo detection methodology to identify advertisements in sport videos. To the best of our knowledge, no work is reported in the literature that explores the possibility of identifying candidate spaces in outdoor scenes for advertisement implantation. In this paper, we attempt to bridge this gap by proposing the use of a bespoke neural network specifically designed for this task of candidate space identification in outdoor scenes.

\subsection{Deep-learning based approaches for instance segmentation}
In our previous work~\cite{dev2018case}, we released the first large scale dataset of outdoor scenes along with manually annotated binary maps of candidate spaces. The annotators manually identified a single candidate space in an image scene, that achieves the \emph{best} aesthetic value according to their subjective judgment. 
In this paper, we formulate the task of identifying candidate spaces in an outdoor scene as a semantic segmentation task. Therefore, it is interesting for us to explore various neural nets that are specifically designed for the task of semantic scene segmentation. Using the dataset of input images and corresponding binary maps, we intend to train various deep networks and check their corresponding efficacy in identifying candidate spaces. The Fully Convolutional Network (FCN) proposed by Long et al.\ is a popular approach of using fully connected layers for semantic segmentation~\cite{long2015fully}. The U-Net model~\cite{ronneberger2015u} proposed by Ronneberger et al.\ has also gained popularity for image segmentation tasks in various application domains. The U-Net has a symmetrical structure, and uses a concatenation operator in the skip connections between the contracting- and the expanding- paths. Recently, for parsing outdoor scenes, the pyramid scene parsing network (PSPNet)~\cite{zhao2017pyramid} has produced satisfying segmentation masks in the cityscapes dataset~\cite{cordts2016cityscapes}. Furthermore, Paszke et al.\ in ~\cite{paszke2016enet} has proposed a deep neural network architecture (ENet) that has the ability to produce satisfactory segmentation masks in low-powered mobile devices. 
More discussion on the performance of these algorithms in identifying candidate spaces is discussed in Section~\ref{sec:obj-eval}. 

\subsection{Proposed model for advert implantation}

In this section, we propose a light-weight convolutional neural network that is specifically designed for the task of identifying candidate spaces. We refer our proposed architecture as \emph{DeepAds} that can assist in proposing candidate spaces for advert integration. Our architecture is inspired from FCN network~\cite{long2015fully}, and is a minimalist version of a large scale encoder-decoder architecture.

\begin{figure}[htb]
  \begin{center}
    \includegraphics[width=0.45\textwidth]{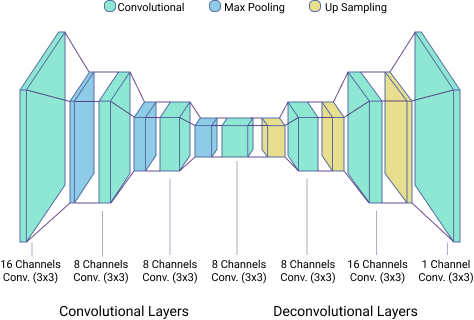}
  \end{center}
  \caption{Our proposed DeepAds model is based on a simple encoder-decoder architecture. The convolutional, max pooling and upsampling layers are color coded for ease of interpretation.}
  \label{fig:deepads-arch}
\end{figure}

Figure~\ref{fig:deepads-arch} describes our proposed architecture. We use regular RGB images as input images with a fixed dimension of $200\times200$. The encoder component of DeepAds gradually reduces the spatial dimension of the input images. The first convolutional layer consists of $16$ channels. The next three convolutional layer consists of $8$ channels respectively. Each filters in these channels are of dimension $3\times3$. The max-pooling layers are inserted between the convolutional layers, as described in Fig.~\ref{fig:deepads-arch}. The decoder component of the DeepAds model recovers the spatial component of the encoded image by gradual upsampling. The first two convolutional layers in the deconvolutional layers consist of $8$ and $16$ channels respectively, each with a filter dimension of $3\times3$. The last convolutional layer consists of a single channel, with $3\times3$ filter size. The output of the DeepAds model is a probabilistic map that provides a degree of likelihood to each pixels of the input image, of integrating a new advertisement in that position. 

\section{Experiments and Results}

\subsection{Dataset}

\begin{figure}[htb]
\centering
\includegraphics[height=0.07\textwidth]{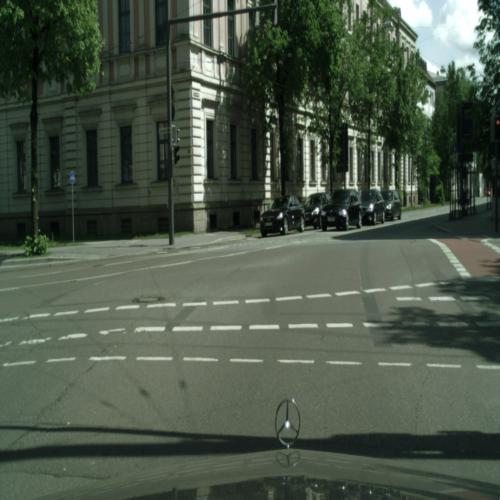}
\includegraphics[height=0.07\textwidth]{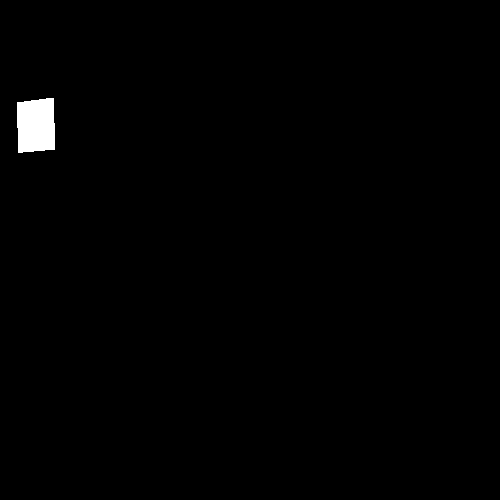}
\includegraphics[height=0.07\textwidth]{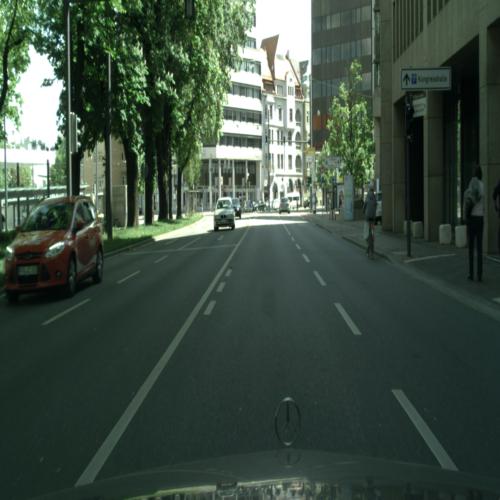}
\includegraphics[height=0.07\textwidth]{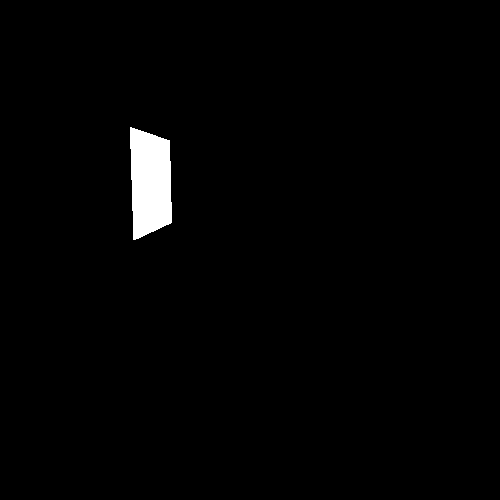}
\includegraphics[height=0.07\textwidth]{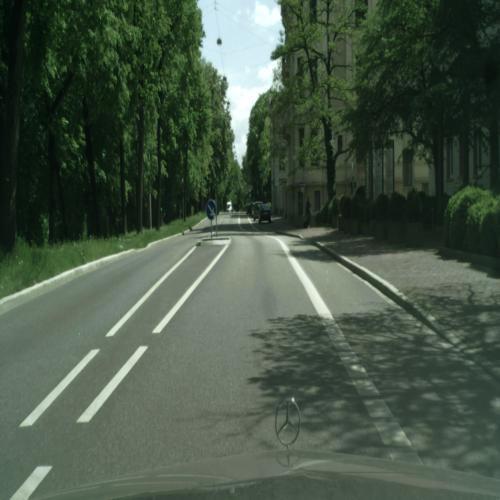}
\includegraphics[height=0.07\textwidth]{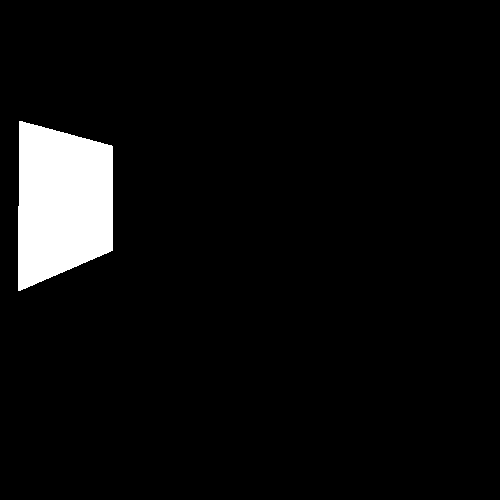}\\
\includegraphics[height=0.07\textwidth]{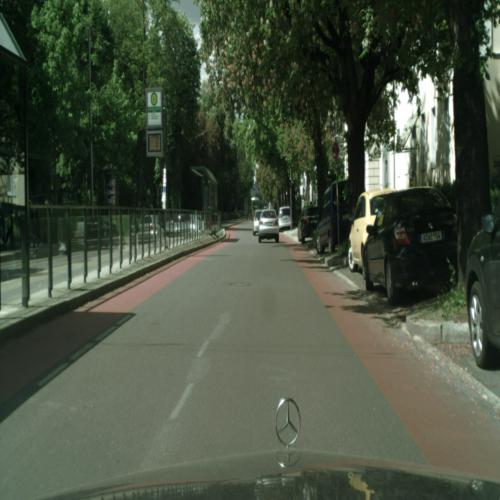}
\includegraphics[height=0.07\textwidth]{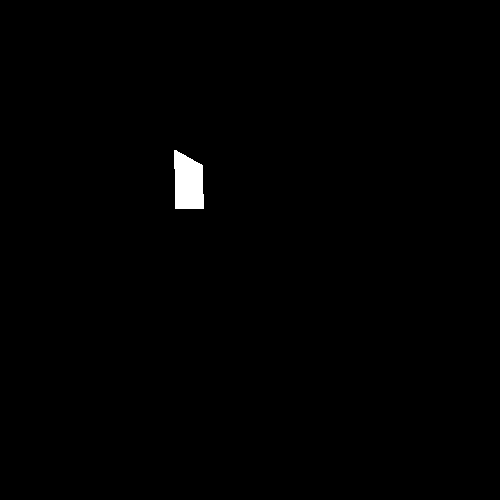}
\includegraphics[height=0.07\textwidth]{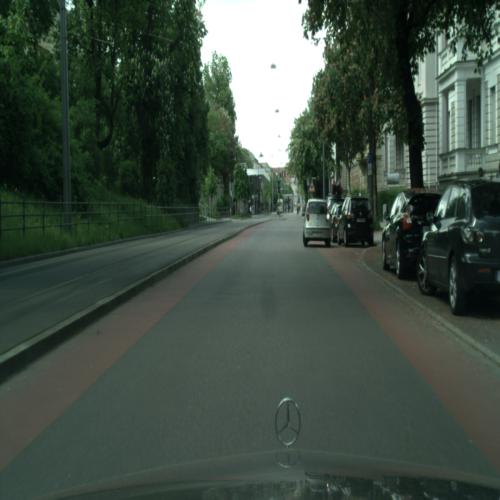}
\includegraphics[height=0.07\textwidth]{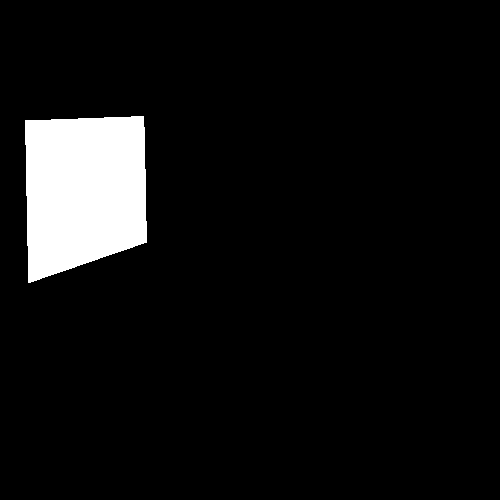}
\includegraphics[height=0.07\textwidth]{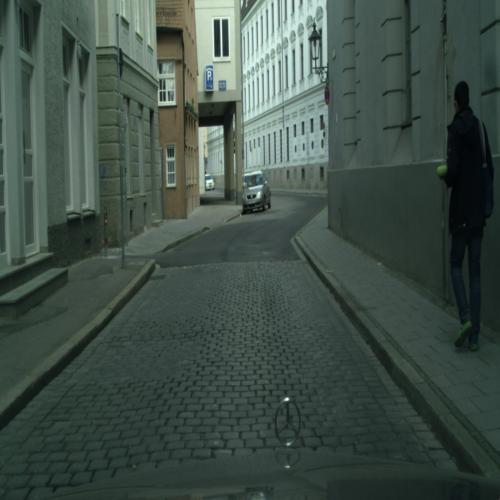}
\includegraphics[height=0.07\textwidth]{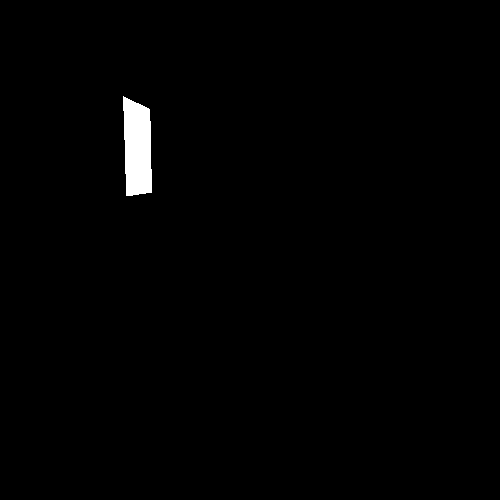}\\
\vspace{0.3cm}
\caption{Representative images of the candidate space dataset. We show the input images and the corresponding binary maps that indicate the best location for integrating a new advertisement.}
\label{fig:wall-images}
\end{figure}

Identifying the \emph{best} candidate location for integrating a new advertisement is a subjective task. Recently, in ~\cite{dev2018case}, we released the first large-scale dataset of outdoor scenes with manually annotated binary candidate maps. We hired paid volunteers to go through a batch of outdoor scenes, and the annotators manually identified the four corners of a candidate space in the scene. We restricted them to identify a single instance of such space. This is intentionally performed such that the volunteers can identify the \emph{best} space, according to their subjective judgement. Figure~\ref{fig:wall-images} shows a few of the representative images of the outdoor scene. The candidate spaces are represented as binary maps, wherein the white region indicate the best candidate space for integrating advertisement, and black region indicate the background. 

This dataset with manually annotated candidate locations provides us a good test-bed to benchmark the various semantic segmentation algorithms. We train the various deep neural nets on this dataset of outdoor scene images and the corresponding binary maps. We aim that the different neural networks can learn the semantics of the outdoor scene, and can automatically identify and suggest appropriate locations for advertisement implantation.

\begin{figure*}[p]
\centering
\includegraphics[height=0.16\textwidth]{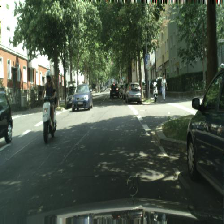}
\includegraphics[height=0.16\textwidth]{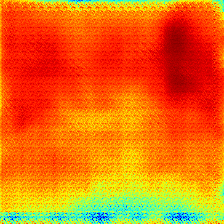}
\includegraphics[height=0.16\textwidth]{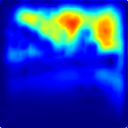}
\includegraphics[height=0.16\textwidth]{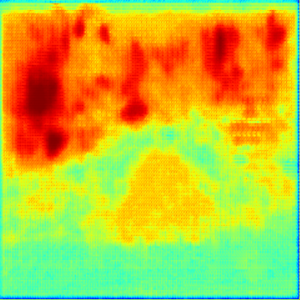}
\includegraphics[height=0.16\textwidth]{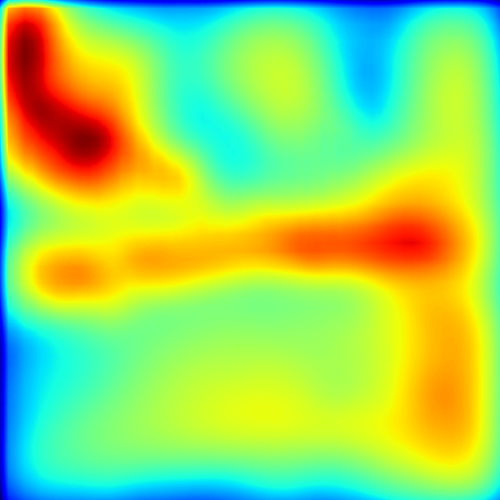}
\includegraphics[height=0.16\textwidth]{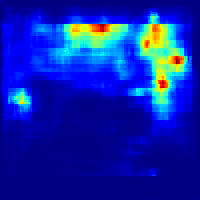}\\
\includegraphics[height=0.16\textwidth]{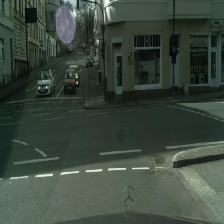}
\includegraphics[height=0.16\textwidth]{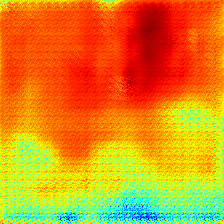}
\includegraphics[height=0.16\textwidth]{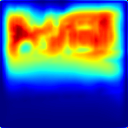}
\includegraphics[height=0.16\textwidth]{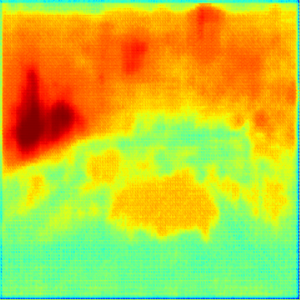}
\includegraphics[height=0.16\textwidth]{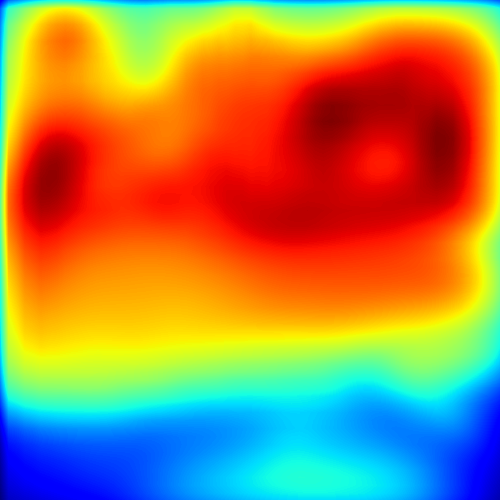}
\includegraphics[height=0.16\textwidth]{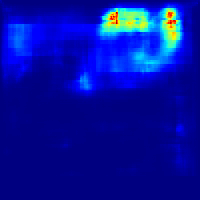}\\
\includegraphics[height=0.16\textwidth]{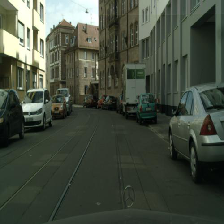}
\includegraphics[height=0.16\textwidth]{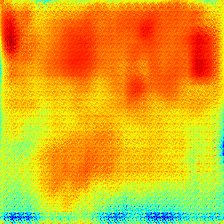}
\includegraphics[height=0.16\textwidth]{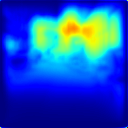}
\includegraphics[height=0.16\textwidth]{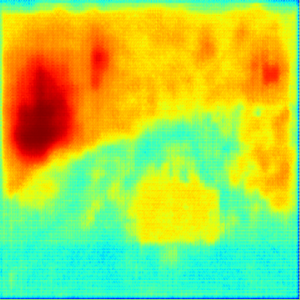}
\includegraphics[height=0.16\textwidth]{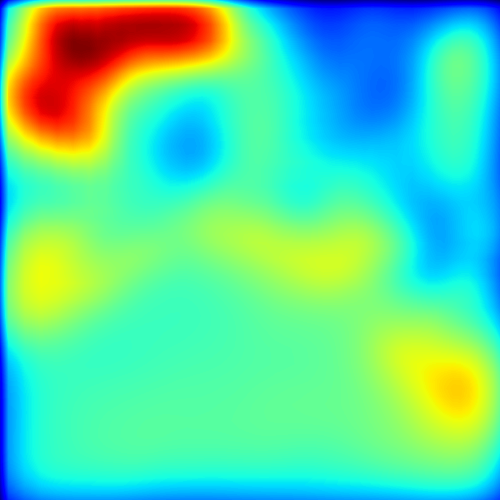}
\includegraphics[height=0.16\textwidth]{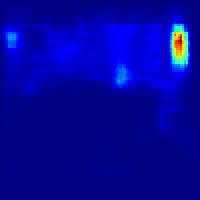}\\
\includegraphics[height=0.16\textwidth]{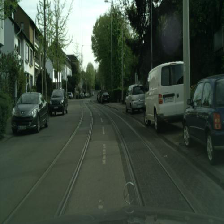}
\includegraphics[height=0.16\textwidth]{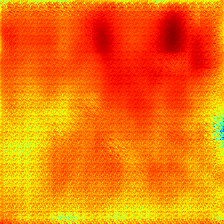}
\includegraphics[height=0.16\textwidth]{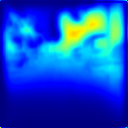}
\includegraphics[height=0.16\textwidth]{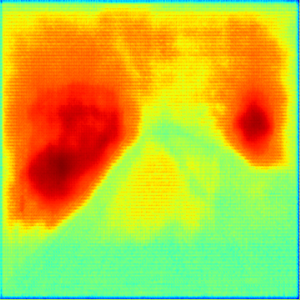}
\includegraphics[height=0.16\textwidth]{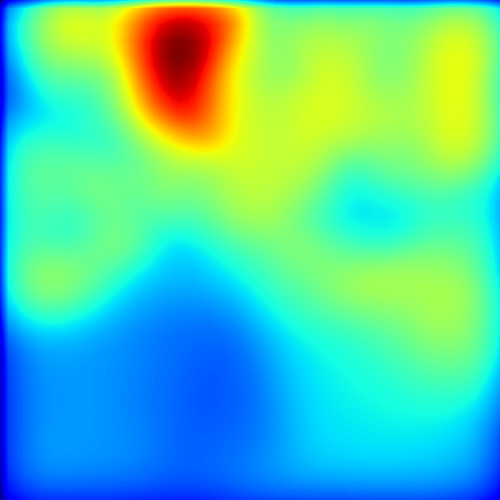}
\includegraphics[height=0.16\textwidth]{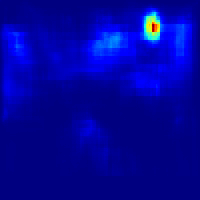}\\
\includegraphics[height=0.16\textwidth]{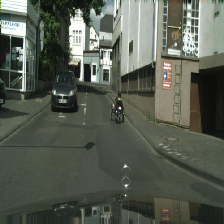}
\includegraphics[height=0.16\textwidth]{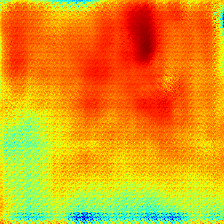}
\includegraphics[height=0.16\textwidth]{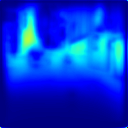}
\includegraphics[height=0.16\textwidth]{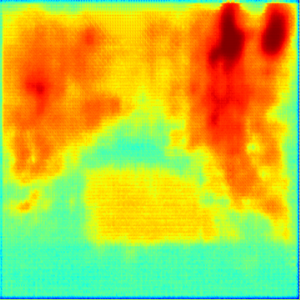}
\includegraphics[height=0.16\textwidth]{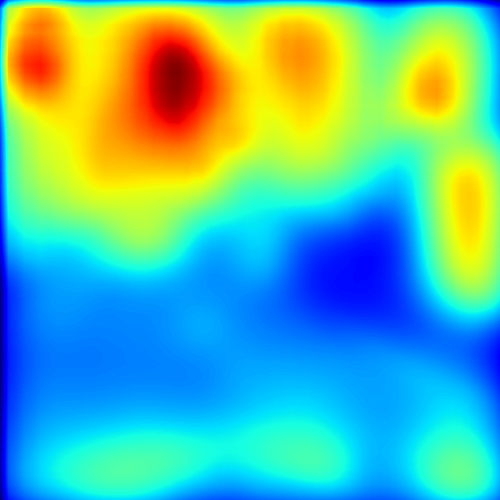}
\includegraphics[height=0.16\textwidth]{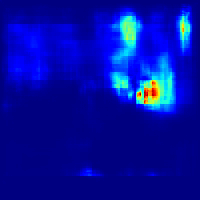}\\
\includegraphics[height=0.16\textwidth]{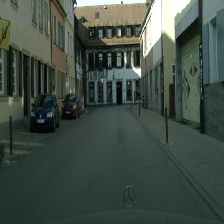}
\includegraphics[height=0.16\textwidth]{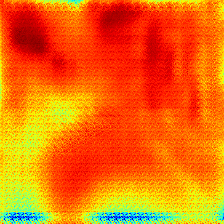}
\includegraphics[height=0.16\textwidth]{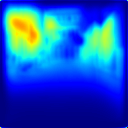}
\includegraphics[height=0.16\textwidth]{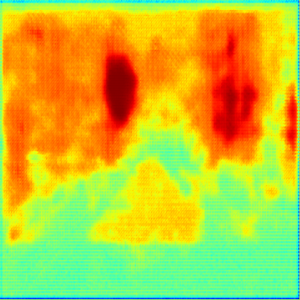}
\includegraphics[height=0.16\textwidth]{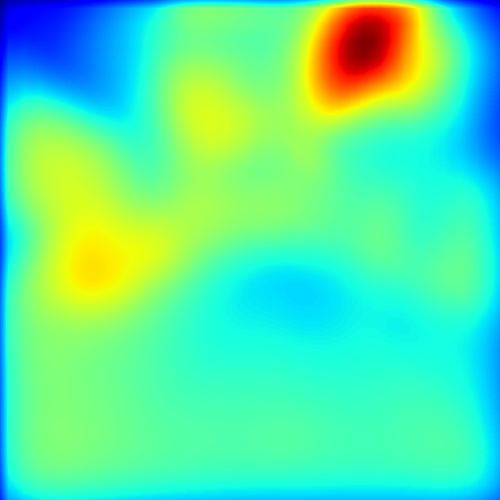}
\includegraphics[height=0.16\textwidth]{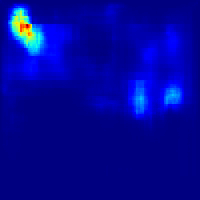}\\
\includegraphics[height=0.16\textwidth]{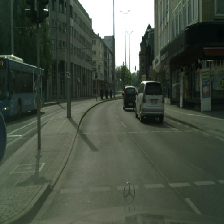}
\includegraphics[height=0.16\textwidth]{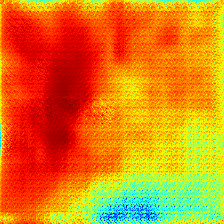}
\includegraphics[height=0.16\textwidth]{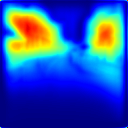}
\includegraphics[height=0.16\textwidth]{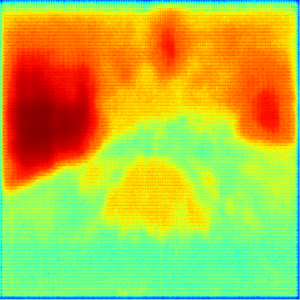}
\includegraphics[height=0.16\textwidth]{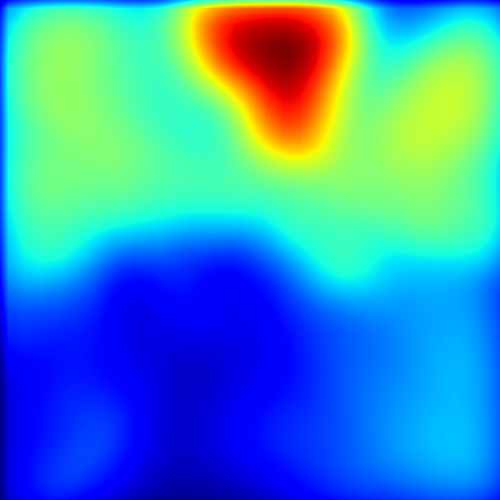}
\includegraphics[height=0.16\textwidth]{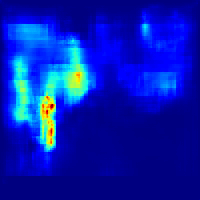}\\
\makebox[0.15\textwidth][c]{Input image}
\makebox[0.15\textwidth][c]{FCN}
\makebox[0.15\textwidth][c]{U-Net}
\makebox[0.15\textwidth][c]{ENet}
\makebox[0.15\textwidth][c]{PSPNet}
\makebox[0.15\textwidth][c]{DeepAds}\\
\vspace{0.3cm}
\caption{Subjective evaluation of several benchmarking algorithms. We represent sample (a) input images, along with probabilistic output map of (b) FCN model, (c) U-Net model, (d) ENet model, (e) PSPNet model, and (f) our proposed DeepAds model.}
\label{fig:subj-eval}
\end{figure*}

\subsection{Subjective Evaluation}
Figure~\ref{fig:subj-eval} shows a few visual illustrations of the different benchmarking approaches, along with our proposed model. We train the various benchmarking approaches in the dataset of candidate spaces. Instead of the final binary image, we show the probabilistic heat map of the different algorithms. The heat-maps are obtained prior to the final \emph{softmax} layer of the individual neural nets. The analysis of the heat-maps, instead of the binary maps make more sense, as it illustrates the likelihood probability of inserting an advertisement at a particular location of the image scene. We will understand how closely the different neural nets can approximate the subjective human judgement of \emph{best} position of advertisement implantation. We observe from Fig.~\ref{fig:subj-eval} that FCN and ENet suggests that the new advertisement can be implanted almost anywhere in the scene. In comparison, the probabilistic results from U-Net and PSPNet are more localized to particular areas of the outdoor scene. In most of the images, the PSPNet suggests the sky region as possible region for advert implantation. The regions identified from U-Net seem to be more plausible candidates for new advertisements. Finally, the DeepAds results suggest localized regions in the scene as possible candidates. We verify these results with the binary ground-truth maps, in order to check the efficacy of the individual neural nets to approximate its results to the subjective judgement of the annotators. More discussion on this objective evaluation in the next section of the paper.

\subsection{Objective Evaluation}
\label{sec:obj-eval}
We report various metrics related to semantic segmentation, in order to provide an objective evaluation of the various networks. We assume that the total number of pixels in class $i$ that are classified to belong to class $j$ is $n_{ij}$. Also, $n_{cl}$ denote the total number of classes. In our case of identifying advert implantation, we have $2$ classes -- possible image region wherein a new advert can be integrated, and otherwise.  We compute the following metrics: pixel accuracy, mean accuracy, mean intersection over union, and frequency weighted intersection over union. These metrics are variants of classification accuracy and intersection over union (IOU), that are commonly used in such tasks of semantic segmentation. They are defined as follows: Pixel Accuracy = $\frac{\sum_{i}^{} n_{ii}}{\sum_{i}^{} t_{i}}$, Mean Accuracy = $\frac{1}{n_{cl}}\sum_{i}^{}\frac{n_{ii}}{t_i}$, Mean Intersection Over Union = $\frac{1}{n_{cl}}\frac{\sum_{i}^{}n_{ii}}{t_i+\sum_{j}^{}n_{ji}-n_{ii}}$, and Frequency Weighted Intersection Over Union = $\frac{1}{\sum_{k}^{}t_k}\frac{\sum_{i}^{}t_in_{ii}}{t_i+\sum_{j}^{}n_{ji}-n_{ii}}$.

Table~\ref{table:result} reports the results of the various benchmarking approaches in the dataset of candidate spaces. We observe that the different metric values correspond well to the subjective evaluation of the different approaches. The PSPNet and U-Net have similar scores in all the metrics. 
The FCN network that has a similar architecture to our proposed neural net performs better. Our light-weight architecture DeepAds performs the best with respect to mean IOU. The encoder-decoder architecture of DeepAds can successfully learn the semantics of the image scene, and can propose localized candidate spaces that are closest to the human subjective judgment. 

\begin{table}[htb]
\small 
\centering 
\begin{tabular}{l|p{1.3cm}|p{1.3cm}|p{1.3cm}|p{1.3cm}}
       & \textbf{Pixel Accuracy} & \textbf{Mean Accuracy} & \textbf{Mean IOU} & \textbf{Frequency Weighted IOU} \\ \hline
FCN    & 0.978 & 0.509 &  0.498  &  0.959  \\ 
PSPNet & 0.545 & 0.625 & 0.284 & 0.529 \\ 
U-Net & 0.619 & 0.727 &   0.327 &  0.601 \\  
Enet &  0.336  &  0.653  &  0.176 & 0.317 \\  
DeepAds & 0.970 & 0.569 & 0.518 & 0.948 \\ \hline
\end{tabular}
\vspace{0.5cm}
\caption{Performance evaluation of the proposed DeepAds model in localizing adverts. }
\label{table:result}
\vspace{-0.2cm}
\end{table}

\subsection{Receiver Operating Characteristics (ROC) curve}
\label{sec:which-value}

As a final comparison amongst the benchmarking methods, we plot the receiver operating characteristics of the various segmentation neural nets. In our binary classification problem of identifying candidate spaces, we assume that $TP$, $TN$, $FP$ and $FN$ denote the true positive, true negative, false positive, and false negative samples respectively. The ROC curve is a plot between False Positive Rate (FPR) and True Positive Rate (TPR). The FPR and TPR are defined as follows: $\mbox{FPR} = \frac{FP}{FP+TN}$ and $\mbox{TPR} = \frac{TP}{TP+FN}$ respectively.

\begin{figure}[htb]
\centering
\includegraphics[width=0.42\textwidth]{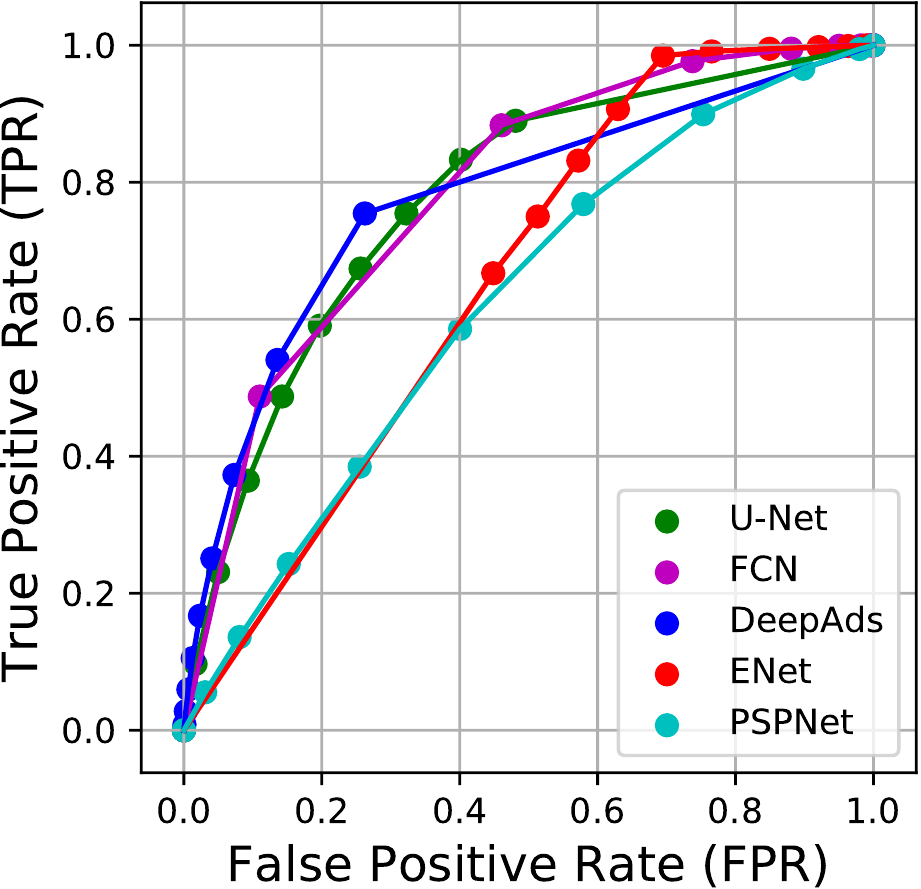}\\
\vspace{0.5cm}
\caption{Receiver Operating Characteristics (ROC) curve of the various benchmarking algorithms across different discrimination thresholds. We observe that our proposed DeepAds model has the largest area under its curve, indicating a competitive performance across the various thresholds.}
\label{fig:roc}
\end{figure}

Figure~\ref{fig:roc} illustrates the receiver operating characteristics curve of the various benchmarking methods. We vary the discrimination threshold and generate the corresponding binary maps from the probabilistic masks. Thereby, we compute the corresponding FPR and TPR values, corresponding to the output binary map and the binary ground-truth map. Each point in the curve in Fig.~\ref{fig:roc} denotes the mean FPR and TPR values averaged over all the testing images. 
We observe that the ENet and PSPNet perform poorly for most of the discrimination thresholds. The FCN and the DeepAds model have a similar performance for most of the threshold values. The U-Net model performs slightly worse than that of FCN model. Our proposed model DeepAds has the largest area under ROC curve indicating a good performance in identifying the candidate spaces in the image scene. 

\section{Conclusion and Future Work}

In this paper, we presented our method of detecting candidate spaces in image- and video- frames for advertisement placements. These spaces are suitable for seamless insertion of advertisements without affecting the flow of the video scene. Our method provides such information as a heat-map prediction based on the input image frame. 
We compared our method with state-of-the-art segmentation networks such as FCN, PSPNet, U-Net and ENet. We showed that our bespoke light-weight architecture surpassed the alternative architectures in producing a successful solution for the task of identifying spaces for virtual advertisements. 

In the future, we are interested to extend our analysis on identifying multiple locations of candidate spaces. 
We intend to maximise the goal of visibility and aesthetic value so that the new implanted advert fall in the visual saliency region of the viewer. We are also interested in looking at handling indoor images and video scenes, in addition to outdoor scenes. It is of prime importance these days, with the introduction of popular sitcoms and user-generated television shows.

\balance 

\bibliographystyle{IEEEbib}

\end{document}